\documentclass[journal]{IEEEtran}
\ifCLASSINFOpdf
\else
\fi
\usepackage{soul}

\usepackage{times}
\usepackage{epsfig}
\usepackage{graphicx}
\usepackage{amsmath}
\usepackage{amssymb}
\usepackage{algorithm}
\usepackage{algorithmic}
\usepackage{mathrsfs}
\usepackage{multirow}

\usepackage{amsthm}
\theoremstyle{plain}% default

\usepackage[section]{placeins}

\usepackage[]{animate}

\usepackage{graphicx}
\usepackage{animate}

\usepackage{float}

\usepackage{multirow}
\usepackage{rotating}
\usepackage{wrapfig}
\usepackage{lineno}
\usepackage{epsfig}
\usepackage{amsmath,amssymb}
\usepackage{color}
\usepackage{booktabs}       % professional-quality tables
\usepackage{amsfonts,amsmath}       % blackboard math symbols
\usepackage{nicefrac}       % compact symbols for 1/2, etc.
\usepackage{microtype} 
\usepackage{float}
\usepackage{amsmath}
 
\usepackage{enumerate}
\usepackage[pagebackref=true,breaklinks=true,letterpaper=true,colorlinks,bookmarks=false]{hyperref}

\begin{document}
%\title{Classification of Patients with ALS from Healthy Controls from Structural MRI via Successive Subspace Learning}
\title{{Interpreting Depression from Question-wise Long-term Video Recording of SDS Evaluation}}

\author{Wanqing Xie$^{\dag}$, Lizhong Liang$^{\dag}$, Yao Lu, Chen Wang, Jihong Shen, Hui Luo*, Xiaofeng Liu*,~\IEEEmembership{Member,~IEEE} % <-this % stops a space
\thanks{W. Xie is with the College of the Mathematical Sciences, Harbin Engineering University, Harbin, China.}
% <-this % stops a space
\thanks{L. Liang is with the School of Computer Science and Engineering, Sun Yat-sen University, Guangzhou 510006, China, and the Affiliated Hospital of Guangdong Medical University, Zhanjiang, Guangdong, 524023, China.}
\thanks{Y. Lu is with the School of Computer Science and Engineering, Sun Yat-sen University, Guangzhou 510006, China.}
\thanks{C. Wang is with the College of the Mathematical Sciences, Harbin Engineering University, Harbin, China.}
\thanks{J. Shen is with the College of the Mathematical Sciences, Harbin Engineering University, Harbin, China.}
\thanks{H. Luo is with the Southern Marine Science and Engineering Guangdong Laboratory (Zhanjiang), Zhanjiang, Guangdong, 524023 and the Marine Biomedical Research Institute of Guangdong, Zhanjiang, 524023, China.}
\thanks{X. Liu is with the Dept. of Neurology, Beth Israel Deaconess Medical Center, Harvard Medical School, Boston, MA, USA, and Suzhou Fanhan Information Technology Co., Ltd, Suzhou, China. (xliu61@mgh.harvard.edu)}

\thanks{$\dag$W. Xie and L. Liang contribute equally. *H. Luo and X. Liu are the corresponding authors.}

\thanks{Manuscript received April 06, 2021; revised May 30, 2021 and June 15, 2021; Accepted June 23, 2021.}}


\markboth{Accepted to IEEE Journal of Biomedical and Health Informatics,~Vol.~14, No.~8, August~2015}%
{Shell \MakeLowercase{\textit{et al.}}: Bare Demo of IEEEtran.cls for IEEE Journals}



\maketitle

\begin{abstract}

Self-Rating Depression Scale (SDS) questionnaire has frequently been used for efficient depression preliminary screening. However, the uncontrollable self-administered measure can be easily affected by insouciantly or deceptively answering, and producing the different results with the clinician-administered Hamilton Depression Rating Scale (HDRS) and the final diagnosis. Clinically, facial expression (FE) and actions play a vital role in clinician-administered evaluation, while FE and action are underexplored for self-administered evaluations. In this work, we collect a novel dataset of 200 subjects to evidence the validity of self-rating questionnaires with their corresponding question-wise video recording. To automatically interpret depression from the SDS evaluation and the paired video, we propose an end-to-end hierarchical framework for the long-term variable-length video, which is also conditioned on the questionnaire results and the answering time. Specifically, we resort to a hierarchical model which utilizes a 3D CNN for local temporal pattern exploration and a redundancy-aware self-attention (RAS) scheme for question-wise global feature aggregation. Targeting for the redundant long-term FE video processing, our RAS is able to effectively exploit the correlations of each video clip within a question set to emphasize the discriminative information and eliminate the redundancy based on feature pair-wise affinity. Then, the question-wise video feature is concatenated with the questionnaire scores for final depression detection.  Our thorough evaluations also show the validity of fusing SDS evaluation and its video recording, and the superiority of our framework to the conventional state-of-the-art temporal modeling methods.

\end{abstract}


% \fi
\IEEEpeerreviewmaketitle

\section{Introduction}

\begin{figure}[t]
\begin{center}
\includegraphics[width=1\linewidth]{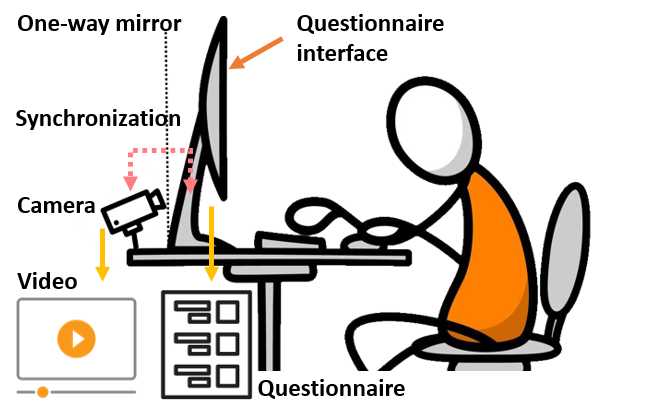}
\end{center} 
\caption{The illustration of collecting SDS questionnaires and the corresponding synchronized face videos. The depression can be better detected with both SDS score and facial expression videos.}
\label{fig1}
\end{figure}

\IEEEPARstart{D}{epression}, a.k.a. major depressive disorder, is a common and serious mental health disorder, while it can be treated \cite{beck2014depression}. Effective early diagnosis can be beneficial for intervention therapy. The earlier that treatment can begin, the more effective it is \cite{beck2009depression}. However, the comprehensive clinical interview for final diagnosis, i.e., clinical golden standard \cite{paykel1985clinical}, can be costly for the large population screening.

Depression is even more prevalent during the COVID-19 pandemic \cite{ettman2020prevalence,hyland2020anxiety}, while the in-person clinician interview can be inconvenient or even prohibitive.

The Self-Rating Depression Scale (SDS) \cite{zung1965self1} is a widely adopted self administrative fast screening questionnaire with twenty questions, which involves affective, psychological, and somatic symptoms related to depression\footnote{\url{https://www.who.int/substance_abuse/research_tools/en/english_zung.pdf}}. Each question is framed in terms of positive and negative statements, and be scored on a Likert scale ranging from 1 to 4. The final result is the sum of each question. We note that the larger score usually indicates the subject is more likely to be a depression patient. Conventionally, we set 50 as the threshold of normal or depression \cite{zung1965self}. The SDS ratings have indicative depression level ranges that may help health assessment and testing \cite{biggs1978validity}.

The ratings include suggestive depression level ranges that may help therapeutic and scientific research, but the SDS outcome can vary from the clinical interview for verifying a depression diagnosis \cite{gabrys1985reliability}. A reason can be the uncontrollable self-administered measure can be easily affected by insouciantly or deceptively answering \cite{zung1967factors}, and producing the different results with the clinician-administered interview, e.g., Hamilton Depression Rating Scale (HDRS) \cite{williams1988structured}. 

\begin{figure*}[t]
\begin{center}
\includegraphics[width=1\linewidth]{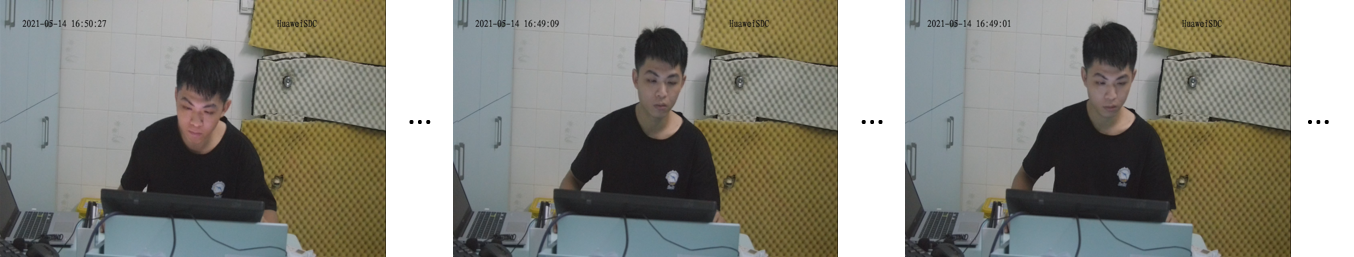}
\end{center} 
\caption{{Illustration of a subject answering a question in SDS. A Software-Defined Camera is adopted with the parameters of 25 fps and 3840$\times$2160 resolution.}}
\label{figexp}
\end{figure*}

Clinically, facial expression (FE) \cite{corneanu2016survey} and the actions \cite{pampouchidou2017automatic,zhu2020improved} can play an important role in clinician-administered evaluation, while FE and actions are underexplored for self-administered evaluation. Actually, expression and actions can be expressive features for many psychiatry analyses \cite{rubinow1992impaired,krause1989facial}. Based on this insight, in this work, we collect a novel dataset of 200 subjects to evidence the validity of self-rating questionnaires with video recording. To provide a more fine-grained connection between the questionnaire and the video, we adopt a Software-Defined Camera (SDC) system which synchronized with the questionnaire software to record the video start from the question showing and end on the score is chosen. For each subject, there are 20 question and video pairs. To extract the region of interests (ROI), the face detector is applied, and the face box is extended 100\% to incorporate the hand action, e.g., head-scratching, chin-touching. Moreover, the answering time may also affect the depression diagnosis \cite{lewinsohn1969depression}.

The video provides additional information for analysis, while it also introduces several challenges for automatical analysis. Firstly, the video of each question can be quite long (e.g., 525 frames in our dataset), and the useful information can be sparse in a long-term sequence. 2) Secondly, the length of each video varies from 50 to 525 frames, which is depends on the question and the participants. 3) Moreover, a practical information fusion scheme is necessary to explore both the SDS evaluation and its question-wise video recording.

To automatically interpret depression from the SDS evaluation and its corresponding FE and action video recording, a redundancy-aware conditional self-attention framework is proposed. Specifically, we resort to a hierarchical model which utilizes a 3D convolutional neural network (CNN) \cite{ji20123d} for local temporal pattern exploration and a self-attention scheme for question-wise global feature aggregation. We factorize the question-wise video into the fix-length clips according to the time of human facial expression developing. With the fix-length input, the 3D CNN is able to extract the local temporal cues efficiently. Then, the clip-wise representation is feed forwarded to a parametric redundancy-aware self-attention (RAS) scheme to eliminate the uninformative signals and extract the representative question-wise feature.

Targeting for the redundant long-term FE and action video processing, our RAS is able to effectively exploit the correlations of each video clip in a question to emphasize the discriminative information and eliminate the redundancy. It traverses the clips within a question to produce a refined representation based on pair-wise feature affinity \cite{wang2018non}. Our residual attention term is able to prioritize discriminative clips while ignoring inferior ones for explicit redundancy reduction. In addition, the temporal sequence is explicitly considered with a Gaussian similarity kernel.

Then, the question-wise video feature is concatenated with the questionnaire scores for final depression detection. Our thorough evaluations also show the validity of fusing SDS evaluation and its video recording, and the superiority of our framework to the conventional state-of-the-art temporal modeling methods.

The main contribution of this work can be summarized as follows:

$\bullet$ To the best of our knowledge, this is the first attempt at exploring depression with both SDS evaluation and its corresponding question-wise face video recording. An elaborate synchronized system is designed, and the final clinician interview results are collected. 

$\bullet$ A practical hierarchical conditional self-attention framework is proposed to explore the long-term variable-length video with 3D CNN for local temporal modeling, redundancy-aware RAS for global attention modeling, and SDS-score conditional question-wise fusion. 

$\bullet$ Our parametric redundancy-aware self-attention (RAS) scheme explicitly emphasizes the discriminative clips and reduces the redundancy based on feature pair-wise affinity, and is aware of the temporal sequence with a Gaussian similarity kernel.

$\bullet$ The systematic and thorough comparisons with the previous temporal modeling methods provide further insights into the potential benefits of our framework. We note that the proposed framework can potentially be generalized to other classification tasks using both questionnaire and video modalities.

\section{Related work}

\noindent \textbf{Deep learning for depression detection} 

In recent decays, there are numerous works dedicated to collect better quality, and larger quantities of depression datasets \cite{pampouchidou2017automatic}. There has been a long history of using SDS evaluation for self-report \cite{zung1965self}. In addition, the facial expression \cite{scherer2013automatic}, eye movement \cite{zhu2020improved} and body action \cite{joshi2013automated} can be the important modality for depression detection \cite{pampouchidou2017automatic}. However, publicly available SDS and its video recording datasets appropriate for incorporating machine learning methods are missing. To the best of our knowledge, this is the first attempt to automatically explore both the SDS evaluation and the corresponding question-wise video.

Moreover, the ``ground truth" of many datasets are the self report (e.g., DAIC-WOZ \cite{gratch2014distress} and AVEC \cite{valstar2013avec}), which is highly unreliable \cite{pampouchidou2017automatic}. The clinician interview has usually been used for final diagnosis \cite{williams2002patient}, which can be costly for large-scale labeling. In this work, all subjects have taken a more comprehensive clinical interview to collect the golden standard label of depression. The scale of our collection, i.e., 200 subjects, is able to support the automatic analysis with the deep learning system.

We note that many modalities can be used for depression detection. For example, \cite{dinkel2019text} targets to the conversation with the Patient Health Questionnaire (PHQ)-8 metric \cite{kroenke2009phq}. \cite{haque2018measuring} propose to fuse the spoken language and 3D facial key points in DAIC-WOZ dataset \cite{gratch2014distress}. The audio is used in \cite{dinkel2019depa} with a self-supervised embedding, and the phoneme feature is used in \cite{muzammel2020audvowelconsnet}. These works try to mimic the clinician-based interview. More recently, the electroencephalography and paralinguistic behaviors are fused with the classifier ensemble for depression detection \cite{zhang2019multimodal}. \cite{shen2020optimal} explores the EEG signal via kernel-target alignment. The fNIRS can also be used for diagnosis \cite{zheng2020feature}. These modalities are able to provide more accurate features, while they are not salable for efficient screening as SDS.

\begin{table}[t]  \caption{The statistics of collected 200 subjects. {M indicates male, while F indicates female.}}\label{tab1}
\centering
\resizebox{0.7\linewidth}{!}{
\begin{tabular}{l|l|l}
\hline\hline
Final diagnosis&	SDS result & Number\\\hline
Normal	& Normal&	86 {(M:42;F:44)}\\ 
Normal	&Depression &20 {(M:13;F:7)} \\ 
Depression &Normal	&20 {(M:11;F:9)}\\ 
Depression&Depression&	74 {(M:38;F:36)}\\\hline\hline
\end{tabular}}
\end{table}

\noindent{\bf{Temporal modeling}} 

Temporal modeling \cite{roddick2002survey} can be the essential part of the video-based classification and analysis tasks, e.g., video facial expression \cite{liu2020identity,liu2021mutual,he2020classification} and action recognition \cite{zhou2017see}. The recursive neural network (RNN) \cite{wang2021automated} is widely used for temporal patterns modeling, which makes dependent sequential processing. However, it is notorious for the long-term forgetting and hard to train \cite{liu2019permutation}. Therefore, its performance can be largely affected if the input is too long. The bi-directional long-short term memory (Bi-LSTM) has been proposed to alleviate the difficulties \cite{schuster1997bidirectional,8121994}, which is still relatively slow in both training and inference. More recently, the 3D CNN \cite{liu2019dependency} is proposed to explore the spatial-temporal patterns in a unified manner. It can be fast processed and has demonstrated good performances in many tasks. Nevertheless, the input to the 3D CNN should be fixed \cite{gao2018revisiting}, limiting its application to the video with variable length, e.g., different questions in our dataset. In this work, we propose to factorize a variable-length video into several fixed-length clips to utilize the 3D CNN to better balance performance and efficiency.

\noindent \textbf{Self-attention and non-local scheme}

Start from the machine translation in \cite{vaswani2017attention}, the attention scheme has demonstrated great potential in many applications. It has been the core block of many successful systems \cite{shaw2018self}. Conventionally, it computes the adjusted output at a position with the weighted sum of all positions in that sentence. A similar philosophy has also been inherited in the non-local algorithms \cite{buades2005non}, which focused on the image denoising task. Pair-wise relationships were also modeled using interaction networks \cite{battaglia2016interaction,hoshen2017vain,watters2017visual,yang2018learning}. Moreover, \cite{wang2018non} proposes a link between self-attention and the wider category of non-local filtering activities. \cite{zhou2017temporal} proposes to learn various time scales of temporal dependencies between video frames. Inspired by these methods, we further adapt this idea to the variable length long-term SDS video analysis.

\begin{figure}[t]
\begin{center}
\includegraphics[width=1\linewidth]{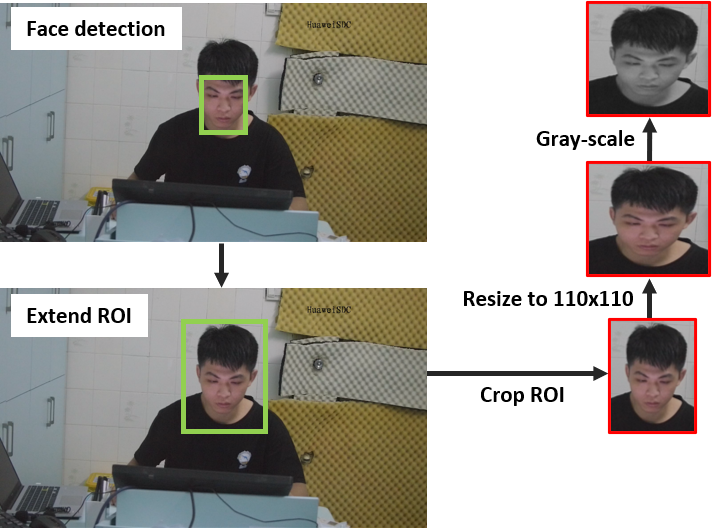}
\end{center} 
\caption{{Illustration of the pre-processing flow, which consists of face detection, ROI extension, ROI crop, resize, and gray-scale processing, sequentially.}}
\label{figexp2}
\end{figure}

\section{Methodology}

\subsection{Data Acquisition and Preprocessing}

We collect the Self-Rating Depression Scale (SDS) questionnaires and the corresponding face video from 200 subjects. The study protocol was approved by the Ethics Committee of the Affiliated Hospital of Guangdong Medical University (No. PJ2021-026).

\begin{figure*}[t]
\begin{center}
\includegraphics[width=1\linewidth]{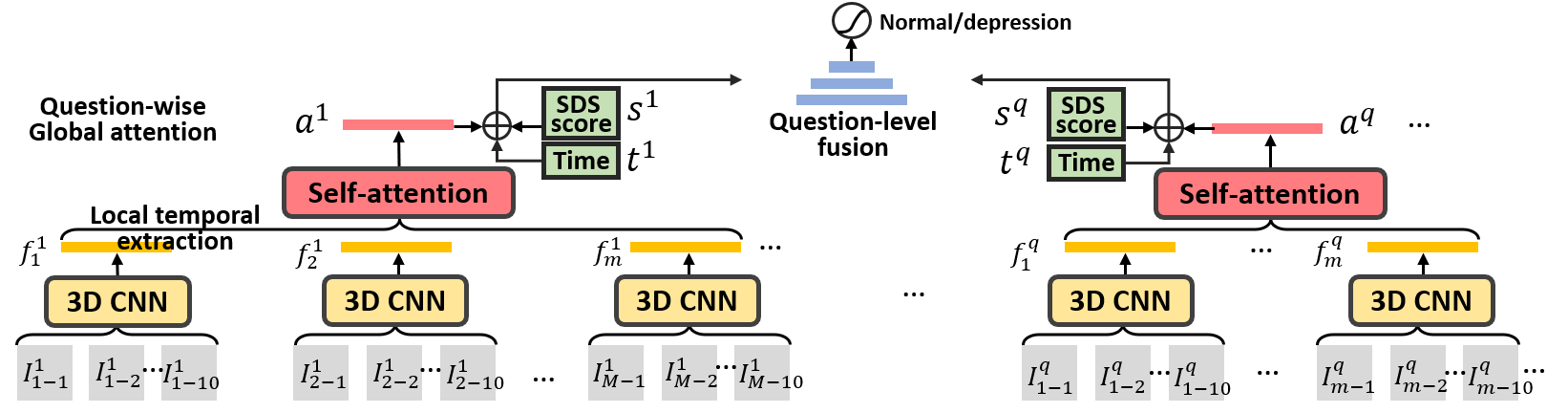}
\end{center} 
\caption{Illustration of the proposed hierarchical conditional self-attention framework, comprising three hierarchically mounted modules for both self-rating questionnaires with video recording, i.e., 3D CNN-based local temporal pattern extraction module, redundancy-aware self-attention module, and question-wise conditional fusion module.}
\label{fig2}
\end{figure*}

Each participant is instructed to sit in a quiet consulting room along and fill the self-report questionnaire following the instruction in the software interface to avoid being affected by the others. Moreover, a Huawei Software-Defined Camera (SDC) is hidden behind a one-way mirror, and the participants do not become aware of the camera in the evaluation. SDC adopts an open-ended software architecture that can flexibly integrate with the code of specific application\footnote{\url{https://e.huawei.com/en/products/intelligent-vision/cameras/software-defined-camera}}. In addition, the data is uploaded to the back-end server for processing. To connect each question with the video, the camera is synchronized with the questionnaire software to record the video start from the question showing and end on the score is chosen. The illustration of our data collection is shown in Fig. \ref{fig1}. In Fig. \ref{figexp}, we provide some frames of a subject for answering a question.

There are 20 questions for SDS evaluation, and they usually take about 10 minutes to complete\cite{zung1965self}. Each participant takes a different time to finish different questions. In our collected dataset, the time of each question varies from 2s to 21s. The 25 fps videos with the resolution of 3840$\times$2160 are collected for each question. Since the region of interest is the face of participants, we use the face detector \cite{zhang2017faceboxes} to crop the face region. To incorporate the hand action of head-scratching \cite{carpenter2000psychotic}, chin-touching \cite{kazdin1985nonverbal}, etc., we extend the face box by 100\% and resize the extended region of interests (ROI) in each frame to 110$\times$110. Moreover, the image is followed by gray processing to reduce the size of the input. The pre-processing flow chart is given in Fig. \ref{figexp2}.

After the self-report SDS evaluation, a more comprehensive clinical interview \cite{paykel1985clinical}, including the clinician-administered Hamilton Depression Rating Scale (HDRS) assessment \cite{williams1988structured}, SCL-90-R Symptom Checklist \cite{derogatis1999scl}, and Self Rating Anxiety Scale (SAS)\cite{jegede1977psychometric}, is taken to help the clinician for confirming a diagnosis of depression. We use the final diagnosis result as our ground truth label.

Considering the self-administrated SDS can be uncontrollable, the result of SDS evaluation can be different from the final diagnosis \cite{gabrys1985reliability}. In Tab. \ref{tab1}, we provide the detailed statistics of the SDS and final results of our 200 subjects. We note that there are only two classes, i.e., normal and depression, in our dataset. We can see that there are about 10\% of subjects have different results. Specifically, the SDS evaluation of 10 subjects is diagnosed as depression, while the other 10 subjects with the high SDS score are diagnosed as normal with the subsequent clinical interview.

\subsection{Hierarchical conditional framework}
The long-term video recording of SDS evaluation inherits rich emotional information, while it also poses several challenges for processing. First, the long video can be redundant, and only very few (i.e., sparse) frames may indicate the useful cues for depression detection. Second, the length of the video can be varied across different subjects and questions.

Considering that the human expression usually takes 200ms to 500ms, it can be reasonable to segment the video into several fixed-length short clips and explore the local temporal patterns within it. We empirically set each video clip to 10 successive frames in our task. With the fixed length, the 3D CNN can be an efficient module for fast processing \cite{liu2019dependency}.

To avoid splitting an expression, we set an overlap ratio to 0.5, that the first clip of the first question $\{I^1_{1-i}\}_{i=1}^{10}$ is from 1 to 10 frames, and the second clip $\{I^1_{2-i}\}_{i=1}^{10}$ is from 6 to 15 frames, respectively. We use the superscript to denote the question from 1 to 20, and the subscript denotes the images in each clip. 
Therefore, for a $N$-frame video, there can be $M=2\times(N/10)-1$ clips. For a 16 seconds video with 25fps in our dataset, we have 79 clips, and only very few of them are not neutral expressions and do not contribute to complementary information. We extract the 128-dimensional feature of each clip and denote the clip-wise representation of the first question as $\{f^1_{m}\}_{m=1}^{M}$, where $m\in\{1,\cdots,M\}$ index the clips within a question. We note that $M$ can be different for different questions, and we omit the question index for simple notations.

After the clip-wise representations are extracted, the global attention is applied on top of them to extract the 128-dimensional question-wise features $\{a^i\}_{i=1}^{20}$ for 20 questions. To adaptively learn the significance of each clip within a question, we resort to a redundancy-aware self-attention scheme. 

\begin{figure}[t]
\begin{center}
\includegraphics[width=0.9\linewidth]{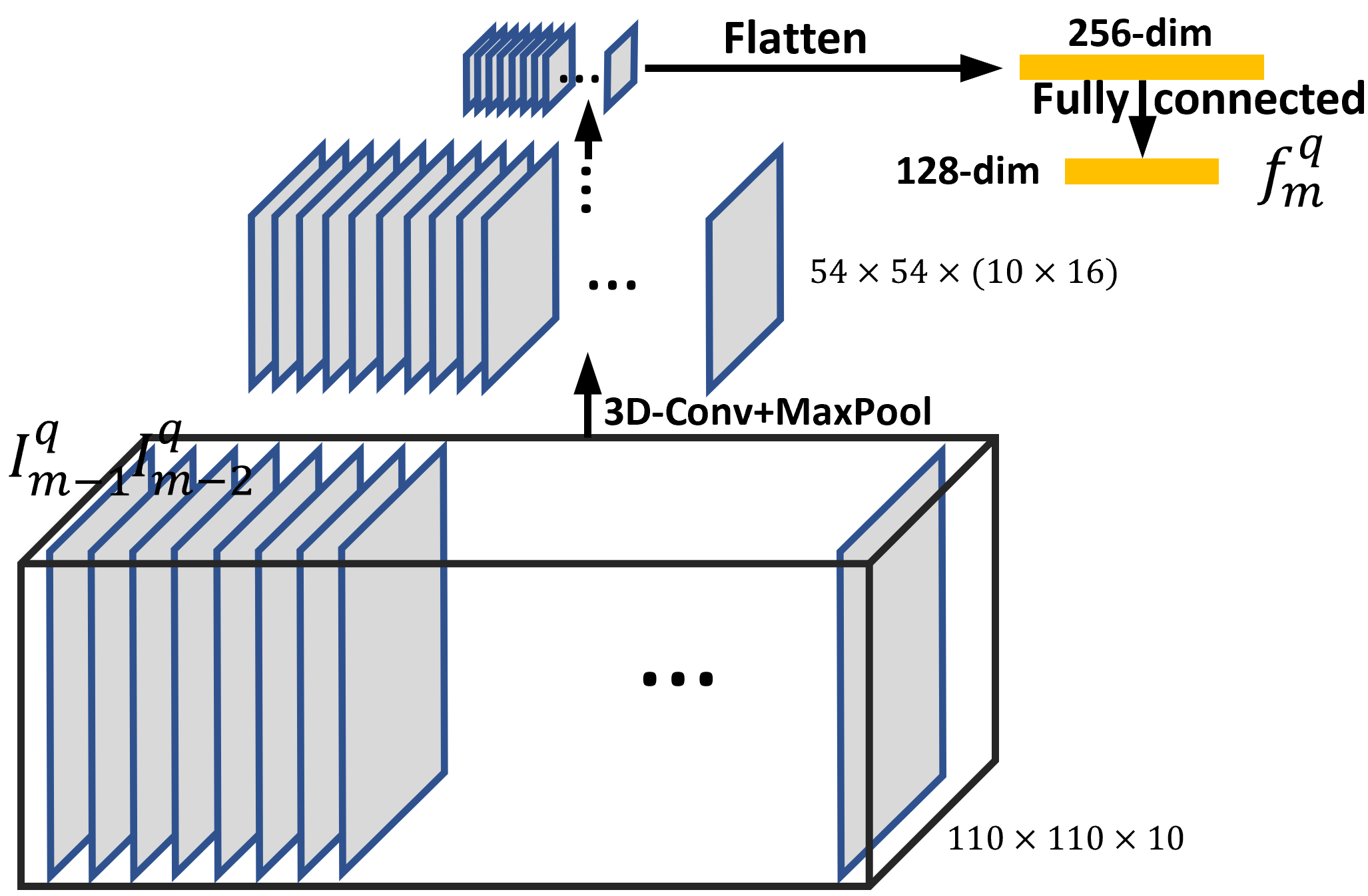}
\end{center} 
\caption{Illustration of the 3D CNN module for local temporal pattern extraction.}
\label{fig3}
\end{figure}

Then, the question-wise features $\{a^i\}_{i=1}^{20}$ are concatenated with the corresponding tabular questionnaire results and the answering time. We note that the SDS has 4 scales, and we use four-dimensional one-hot vector to encode the choice of each problem. The answering time is also concatenated as a one-dimensional scalar. The concatenated feature of each question can be a 133 dimension vector, and all of the 20 questions are concatenated together to form a 2660-dimensional questionnaire and video fused feature. We use the fully connected layers with the sigmoid output unit for binary classification, i.e., normal or depression.

We note that all of the 3D CNN and self-attention modules are shared for all of the clips and questions. In the following subsections, we provide the detailed framework of our hierarchically constructed 3D CNN, redundancy-aware attention, and question-level fusion modules.

\subsection{3D CNN for local temporal exploration}

3D CNN has demonstrated its effectiveness of fast temporal representation extraction for relatively short fixed-length video \cite{ji20123d}. We apply the standard 3D convolution operation to model the relationships between the successive frames. We illustrate the basic framework of 3D CNN in Fig. \ref{fig3}. After a few 3D Convolutional and maxpooling layers, we get a 256-dimensional feature vector, which is sent to a fully connected layer and result in a 128-dimensional clip-wise representation $f^i_m$, where $i$ and $m$ index the 20 questions and $M$ clips within this question, respectively. 

Considering the height and width dimensions of the video clips (e.g., 110) is usually much larger than the frame-wise dimension (e.g., 10), the first three maxpooling layers only half the height and width dimensions and denote as (2$\times$2$\times$1)-(1$\times$1$\times$1). In the 4-th maxpooling layer, we half all of the three dimensions and denote as (2$\times$2$\times$2)-(1$\times$1$\times$1). The detailed network structure is given in Tab. \ref{tab2}.

\begin{table}[t]
\caption{The detailed structure of our 3D CNN module for local temporal feature extraction.} \vspace{-5pt}  
\centering % used for centering table
\resizebox{1\columnwidth}{!}{%
\begin{tabular}{l | l | l} % centered columns (4 columns)
\hline\hline %inserts double horizontal lines
Input Size&Type& Filter Shape   \\ [0.5ex] % inserts table
%heading
\hline % inserts single horizontal line

$[110\times110\times10]$&3D-Conv& [16 kernels of $3\times3\times3$]\\
$[108\times108\times(10\times16)]$&MaxPool& (2$\times$2$\times$1)-(1$\times$1$\times$1)\\
\hline

$[54\times54\times(10\times16)]$&3D-Conv& [32 kernels of $3\times3\times3$]\\
$[52\times52\times(10\times32)]$&MaxPool& (2$\times$2$\times$1)-(1$\times$1$\times$1)\\
\hline

$[26\times26\times(10\times32)]$& 3D-Conv& [64 kernels of $3\times3\times3$]\\ 
$[24\times24\times(10\times64)]$&MaxPool& (2$\times$2$\times$1)-(1$\times$1$\times$1)\\
\hline

$[12\times12\times(10\times64)]$& 3D-Conv& [128 kernels of $3\times3\times3$]\\
$[10\times10\times(10\times128)]$&MaxPool& (2$\times$2$\times$2)-(1$\times$1$\times$1)\\
\hline

$[5\times5\times(5\times128)]$& 3D-Conv& [256 kernels of $5\times5\times5$]\\

\hline

$[1\times1\times(1\times256)]$ & Flatten & N/A\\\hline

256-dim &  FC & 128-dim\\

\hline\hline
 
\end{tabular}
\label{tab2} % is used to refer this table in the text
}
\end{table}

\subsection{Redundancy-aware self-attention}

To explore the correlations between each clip, we resort to the affinity of clip-wise feature vectors. We use $i$ to index the $M$ clips within a question, and the $i$-th vector is regarded as a probe vector. $j$ index the other $M-1$ clips other than the $i$-th clip. We note that different questions can have a different number of fix-length clips, while the 3D CNN is not applicable \cite{liu2018dependency,liu2019dependency}. 
   
In our redundancy-aware self-attention module, we configure several self-attention blocks, which are indexed by $l\in\{1,2\cdots,L\}$. $L$ is the total number of stacked sub-self attention blocks. In each self-attention block, we first traverse all of the clips to set the current clip as the probe. Then, we traverse the M-1 clips other than the probe to explore their correlations with the current probe clip.

Specifically, our self-attention block can be formulated as \begin{equation}f_{i}^{q(l)}=f_{i}^{q(l-1)}+\frac{\Omega^{(l)}}{C_{i}}\sum_{\forall {j}}\omega(f_{i}^{q(0)},f_{j}^{q(0)})(f_{j}^{q(l-1)}-f_{i}^{q(l-1)})\Delta_{i,j}\nonumber\label{con:2}\end{equation}\begin{align}{C_{i}}=\sum_{\forall j}\omega(f_{i}^{q(0)},f_{j}^{q(0)})\Delta_{i,j}; l=0,1,\cdots,L \label{111} \end{align}
where ${\Omega^(l)}\in \mathbb{R}^{1\times1\times 128}$ is the weight vector to be learned and ${f}_{i}^{q(0)}={f}_{i}^{q}$. The response is normalized by ${C_{i}}$. The pairwise affinity $\omega(\cdot,\cdot)$ is an scalar.

The operation of $\omega$ in Eq. \eqref{con:2} has many possible function candidates \cite{wang2018non,zhou2017temporal}. We simply choose the embedded Gaussian given by \begin{align}
    \omega(f_{i}^{q(0)},f_{j}^{q(0)})=e^{\psi{(f_{i}^{q(0)})}^T\phi(f_{j}^{q(0)})},
\end{align}
where $\psi({f_{i}^{q(0)})={\Psi}f_{i}^{q(0)}}$ and $\phi(f_{j}^{q(0)})={\Phi}f_{j}^{q(0)}$ are two embedding functions, and $\Psi$, $\Phi\in\mathbb{R}^{128\times 128}$ are the corresponding learnable mapping matrix \cite{wang2018non}.

\begin{figure}[t]
\begin{center}
\includegraphics[width=1\linewidth]{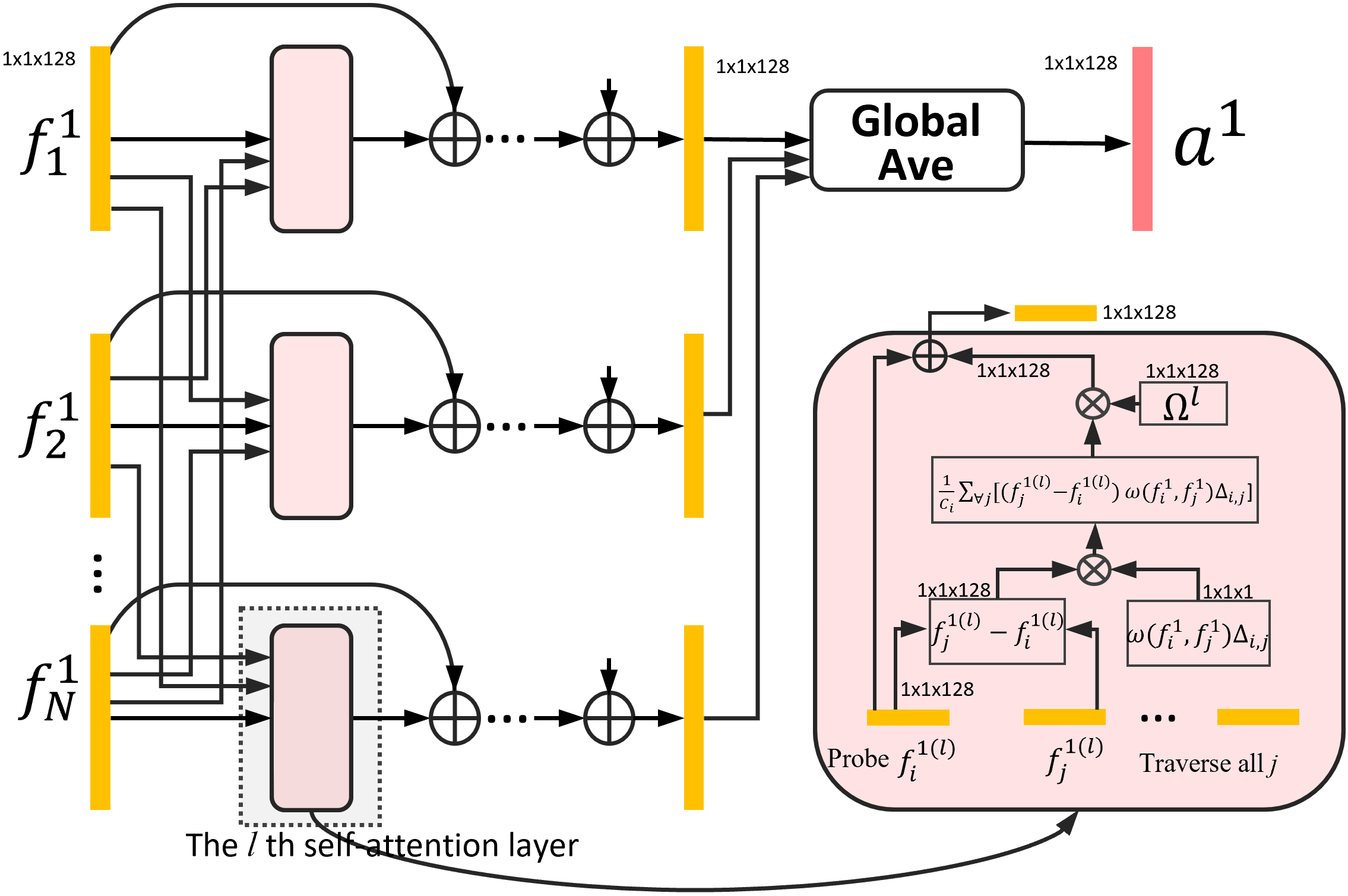}
\end{center} 
\caption{Illustration of the proposed redundancy-aware self-attention module.}
\label{fig4}
\end{figure}

\begin{figure*}[t]
\begin{center}
\includegraphics[width=1\linewidth]{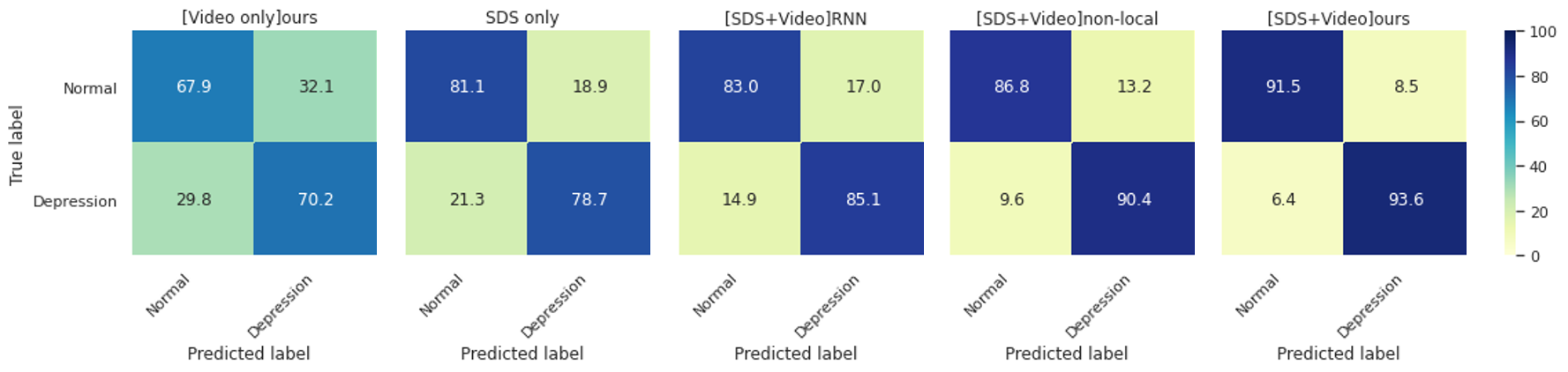}
\end{center} 
\caption{The normalized confusion matrix of the different methods and modalities. From left to right, our hierarchical model using only video, sum operation with tabular SDS evaluation results, SDS and video with RNN, non-local and our hierarchical redundancy-aware self-attention.}
\label{figresult1}
\end{figure*}

The residual term, i.e., $f_{j}^{q(l-1)}-f_{i}^{q(l-1)}$, is the difference of the neighboring feature ($i.e.,f_{j}^{q(l-1)}$) and the current probe feature $f_{i}^{q(l-1)}$. If $f_{j}^{q(l-1)}$ incorporates complementary information or more significant cues compared to the current probe feature $f_{i}^{q(l-1)}$, then our redundency-aware attention scheme will eliminate the information from the inferior $f_{i}^{q(l-1)}$ and emphasis the more discriminative $f_{j}^{q(l-1)}$. Compared to the original non-local network which uses only $f_{j}^{q(l-1)}$ \cite{wang2018non}, our formulation can be more similar to the diffusion maps \cite{tao2018nonlocal}, graph Laplacian \cite{chung1997spectral} and non-local image processing \cite{gilboa2007nonlocal}. All of them are non-local analogues \cite{du2012analysis} of local diffusions, which are expected to be more stable than its original non-local counterpart \cite{wang2018non} due to the nature of its inherit Hilbert-Schmidt operator \cite{du2012analysis}. We not that the the current probe feature will be added back in the last as the residual neural network. Therefore, the previous steps in the self-attention block is adjusting the information of each clip that need to be emphasised and translate to the later blocks.

Since the pair-wise residual term takes all possible clips into consideration and not following the sequential modeling, it does not suffer from long-term forgetting. Therefore, it can be an ideal choice for the attention modeling of many clips. In addition, the weighted average operation is able to take any number of inputs, which is fit for the different clip numbers in different questions. We note that the input and output of a self-attention block have the same size. Specifically, the $M$  inputs with the size of $1\times1\times128$ will be processed to $M$ outputs with the size of $1\times1\times128$.  

Permutation-invariant is a special property of a self-attention scheme \cite{liu2019permutation}. Since we use the sum operation in Eq. \ref{111} to fuse the pair-wise residual terms. In the previous self-attention-based video analysis works, each frame is regarded as independent of each other and discards the sequential patterns \cite{wang2018non}. Our video clips are inherently sequential data, and have an overlap with the neighboring clips. For exploiting the temporal patterns, the Gaussian kernel is proposed for sequential neighboring distance measure \begin{align}
    \Delta_{i,j}={\rm exp}{(\frac{{\parallel m_{i}-m_{j} \parallel}_2^2}{\sigma })},
\end{align}
where $m_{i}$, $m_{j}\in\mathbb{R}$ represent the position of $i^{th}$ and $j^{th}$ feature vectors in the video of a question, respectively. $\sigma$ is a hyperparameter to control the shape of Gaussian Kernel.

After several concatenated self-attention blocks, the global pooling \cite{wang2018non} is added on these $M$ feature maps for element-wise averaging. The final output is a question-wise video feature $a^q\in\mathbb{R}^{1\times1\times128}$.

\begin{table}[t]
\caption{The detailed fully-connected structure of our question-level fusion.} \vspace{-5pt}  
\centering % used for centering table
\resizebox{0.9\columnwidth}{!}{%
\begin{tabular}{l | l | l} % centered columns (4 columns)
\hline\hline %inserts double horizontal lines
Input Size&Type& Filter Shape   \\ [0.5ex] % inserts table
%heading
\hline % inserts single horizontal line

[128+4+1]$\times$20 &Concatenate& 2660 \\
2660 &FC& 1024 with ReLU \\
1024 &FC& 256 with ReLU \\
256 &FC& 1 with sigmoid \\

\hline\hline
 
\end{tabular}
\label{tab3} % is used to refer this table in the text
}
\end{table}

\subsection{Question-level conditional fusion}   
   
The SDS questionnaire score of each question is denoted as $s^q\in\mathbb{R}^4$. Moreover, we empirically found that the answering time of each question can also be helpful for the diagnosis. Therefore, we also concatenate the video length of each question $t^q\in\mathbb{R}$. For a video that takes 3 seconds, we set its $t^q$ to 3. The too short or too long answering maybe unreliable \cite{lewinsohn1969depression}. 

We concatenate $a^q$, $s^q$ and $t^q$ to form a 133 dimension feature vector for each question. Then, the 20 question-wise video and questionnaire features are concatenated to a 2660 dimensional feature for a final depression diagnosis. The fully connected layers are adopted, and the detailed network structure is given in Tab. \ref{tab3}. The widely used Rectified Linear Unit (ReLU) \cite{7280578} is used as the non-linear mapping function between each fully connected layer. 

We use the sigmoid output unit $p=\frac{1}{1+e^{out}}\in(0,1)$ for binary classification, where $out$ is the network prediction scalar in the last layer and will be normalized to a probability value, i.e., the likelihood of this subject is depression patient. We note that we indicate the normal subject with 0 (i.e., $y=0$) and the depression subject with 1 (i.e., $y=1$) according to the final clinician interview. To automatically train our model with backpropagation, we use the binary cross-entropy loss as the optimization objective. 
\begin{align}
   \mathcal{L}= -y\text{log}(p)-(1-y)\text{log}(1-p),
\end{align}
which has the zero loss if $p$ matches its corresponding $y$. In addition, we simply set the threshold of binary prediction to $p=0.5$ for testing.

\section{Experiments}

In this section, we compare the classification performance of our VoxelHop against 3D CNN-based classification. We also provide a systematic ablation study and sensitive analysis to demonstrate the effectiveness of the design choice of our framework.

\subsection{Implementation Details and metrics}

All the experiments were implemented using the widely adopted deep learning library Pytorch \cite{paszke2017automatic} on our server with an NVIDIA V100 GPU, Xeon E5 v4 CPU with 128GB memory. Our model is trained  with the Adam \cite{kingma2014adam} optimizer with the hyper-parameters of $\beta_{1}$=0.9 and $\beta_{2}$=0.999. We used a batch size of 2 for our dataset. The networks of our framework and the compared methods are trained for 200 epochs for a fair comparison. We report the results of five random initialization, and provide the standard deviation ($\pm sd$) along with the average performance. We note that the 3D CNN and the redundancy-aware self-attention modules are shared for all of the questions, which can be processed parallelly. {We empirically set $L=5$}. The training takes about 8 hours, while the average inference time of a subject only takes 1.3s. {The threshold of binary classification testing is set to 0.5.}

We adopt the five-fold cross-validation for the 200 subjects in our dataset. Specifically, we split the dataset into five subsets, and each has 40 subjects. We note that there are no overlap w.r.t. subjects between two folds. Then, we sequentially select a fold as our testing set (i.e., 40 participants), while the remaining four folds (i.e., 160 participants) are used for training.   

For performance evaluation, we use the widely accepted binary classification metrics of accuracy, sensitivity (i.e., recall), specificity. More formally:
\begin{align}
    accuracy&=\frac{TP+TN}{TP+TN+FP+FN},\\
    sensitivity&=\frac{TP}{TP+FN},\\
    specificity&=\frac{TN}{TN+FP},
\end{align}
where TP, TN, FP, and FN indicate true positive, true negative, false positive, and false negative, respectively. We note that the positive and negative corresponding to the depression and normal, respectively. 

In addition, by varying a threshold, the receiver-operating characteristic (ROC) curve plots the true positive rate (TPR) against the false positive rate (FPR). It demonstrates the diagnostic performance of a binary classification algorithm. The larger area under the curve (AUC), the better performance. We note that the TPR$=\frac{TN}{TN+FP}$ and FPR$=\frac{FP}{FP+TN}$.

\begin{table}[t]
\caption{Comparison of the binary classification performance.} \vspace{-5pt}  
\centering % used for centering table
\resizebox{1\columnwidth}{!}{%
\begin{tabular}{l | c | c | c} % centered columns (4 columns)
\hline\hline %inserts double horizontal lines
Methods&Accuracy& Sensitivity & Specificity \\ [0.5ex] % inserts table
%heading
\hline % inserts single horizontal line
$[\text{SDS only}]$sum & 0.800$\pm$0.000 & 0.787$\pm$0.000 & 0.811$\pm$0.000\\
$[\text{Video only}]$ours &  0.690$\pm$0.006   & 0.660$\pm$0.009   &  0.720$\pm$0.007  \\
$[\text{SDS+Video}]$RNN &  0.840$\pm$0.007  &   0.816$\pm$0.007 &   0.846$\pm$0.006  \\
$[\text{SDS+Video}]$non-local &   0.885$\pm$0.005  &  0.859$\pm$0.003  &  0.893$\pm$0.007   \\
$[\text{SDS+Video}]$ours &  \textbf{0.925$\pm$0.004}   & \textbf{0.907$\pm$0.006}   &  \textbf{0.924$\pm$0.005}  \\
\hline\hline
\end{tabular}
\label{resulttab2} % is used to refer this table in the text
}
\end{table}

\subsection{Baselines and Comparison results}

With our SDS and video dataset, there can be three choices of the modality, i.e., SDS only, video only, and both of them. With only the SDS evaluation result, we can simply add the score of 20 questions and using the threshold of 50 for normal and depression classification. According to the statistics in Tab. \ref{tab1}, we can see that the SDS results can be different from the clinician diagnosis. We also tried only using the video modality for classification, which did not concatenate $s^q$ for the question-level fusion.

It is clear that using both SDS and video can outperform SDS only by a large margin, which evidenced the effectiveness of the additional video modality. The complementary information in the corresponding video is helpful to detect depression more accurately. It is also appealingly that even we only use the video, we are also able to predict the depression with an accuracy of 0.69, which is higher than the chance probability of 0.5.

To demonstrate the effectiveness of our model, we also applied two baseline methods, i.e., recurrent neural networks (RNN) \cite{schuster1997bidirectional}, and non-local networks \cite{wang2018non}, for comparison. We note that this is also the first attempt to apply RNN and non-local attention for SDS and video analysis. 

\begin{figure}[t]
\begin{center}
\includegraphics[width=1\linewidth]{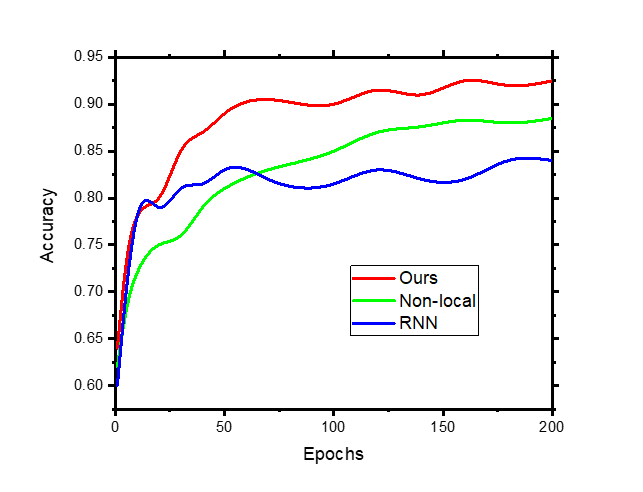}\vspace{-10pt}
\end{center} 
\caption{The accuracy with respect to the number of epochs for different methods.}
\label{figepoch}
\end{figure}

RNN is a typical choice for temporal modeling \cite{pascanu2014construct}. Specifically, we use the bi-directional LSTM \cite{8121994} for our video feature extraction. Moreover, the non-local scheme \cite{wang2018non} is recently proposed to address the long-term forgetting of RNN \cite{pascanu2013difficulty}. We use RNN or non-local to replace the 3D CNN and redundancy-aware self-attention in our framework to extract the 128-dimensional question-wise feature representation. The quantitative evaluation results are shown in Tab. \ref{resulttab2}. In addition, the accuracy of different epochs is plotted in Fig. \ref{figepoch}. Our proposed framework achieves significantly better performance than the RNN and non-local counterparts.

\begin{table}[t]
\caption{Ablation study of the different settings.} \vspace{-5pt}  
\centering % used for centering table
\resizebox{0.8\columnwidth}{!}{%
\begin{tabular}{l | c | c } % centered columns (4 columns)
\hline\hline %inserts double horizontal lines
Methods&Accuracy & AUC \\ [0.5ex] % inserts table
%heading
\hline % inserts single horizontal line

Ours & \textbf{0.925$\pm$0.004}   &  \textbf{0.918$\pm$0.005}  \\\hline
Ours:color &   \textbf{0.925$\pm$0.006} &  {0.917$\pm$0.007}  \\
Ours:non-local &  0.912$\pm$0.005  &  0.904$\pm$0.006   \\
Ours:w/o Time &  0.921$\pm$0.004    &  0.912$\pm$0.004  \\
Ours:w/o$\Delta$ &  0.922$\pm$0.006    &  0.913$\pm$0.006  \\
Ours:$f_{j}^{l-1}$ &   0.920$\pm$0.005   &  0.911$\pm$0.005   \\
Ours:$(\Psi\Phi)^L$ &  0.925$\pm$0.006     &  0.918$\pm$0.007 \\
\hline

{Ours+gender} & {0.924$\pm$0.007}   &   {0.918$\pm$0.006}  \\
{Our:MLP} & {0.864$\pm$0.006}   &   {0.859$\pm$0.005}  \\
{Our:SLF} & {0.905$\pm$0.005}   &   {0.896$\pm$0.007}  \\

\hline\hline
\end{tabular}
\label{resulttab} % is used to refer this table in the text
}
\end{table}

\begin{figure*}[t]
\begin{center}
\includegraphics[width=1\linewidth]{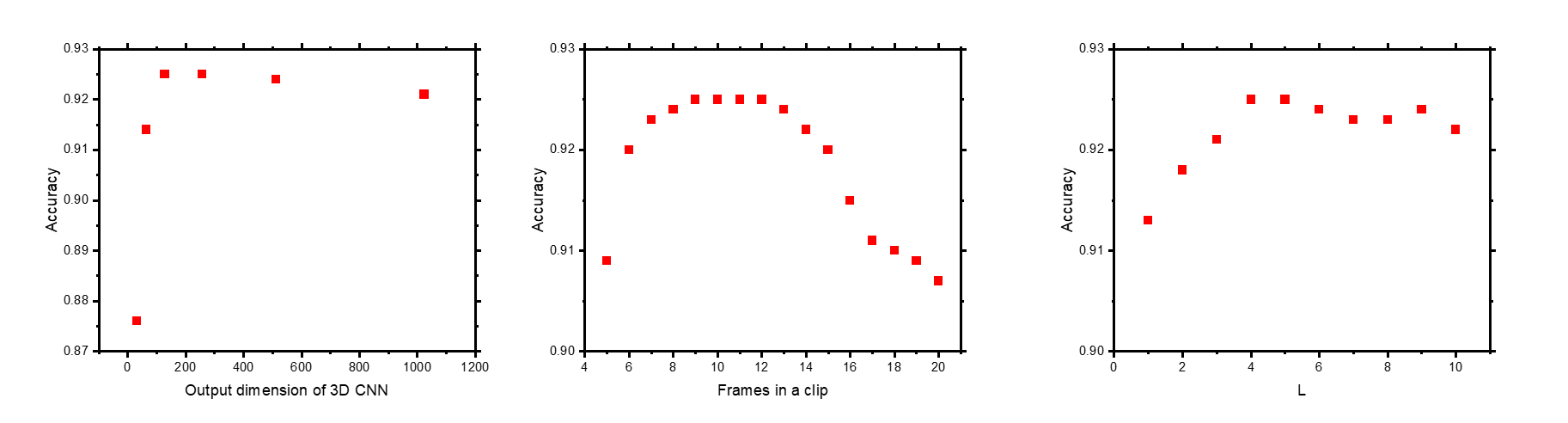}\vspace{-5pt}
\end{center} 
\caption{Sensitivity study of using different output dimension of 3D CNN, length of each clip, and the number of self-attention blocks.}
\label{sensitive}
\end{figure*}

\subsection{Ablation study}

We also provide the systematical ablation study for our framework modules. 

$\bullet$ Our:color indicates using RGB frames as input, which without the gray-scale pre-processing.  We note that we can simply modify the 3D CNN for multi-channel input, while the computation cost can be significantly increased.

$\bullet$ Our:non-local indicates using the conventional non-local \cite{wang2018non} as the alternative of our redundancy-aware self-attention module. We note that the original non-local \cite{wang2018non} is also first introduced to depression detection in this paper, and is regarded as a baseline.

$\bullet$ Our:w/o time denotes the $t^q$ is not concatenated for the question-level fusion. 

$\bullet$ Our:$f_{j}^{l-1}$ refers to using ${\footnotesize f_{j}^{l-1}}$ instead of the difference term  $f_{j}^{q(l-1)}-f_{i}^{q(l-1)}$ in Eq. \eqref{con:2}. It does not explicitly consider the redundancy and lead to lower accuracy. 

$\bullet$ Our:$(\Psi\Phi)^L$ indicates using the embedded Gaussian pair-wise affinity for every block, which has similar performance but usually doubles the training time.  

$\bullet$ Our:w/o$\Delta$ indicates excise $\Delta_{i,j}$ in Eq. \eqref{con:2} and does not taking the temporal sequence into consideration.

{$\bullet$ Our+gender indicates that we add the 1/0 label of male/female along with the time. }

{$\bullet$ Our:MLP indicates that only using MLP to fuse the score and SDS time.}

{$\bullet$ Our:SLF indicates that using score-level fusion instead of feature level fusion.}

The results are provided in Tab. \ref{resulttab}. The relatively inferior performance of the compared settings demonstrates the effectiveness of our choices. {By adding the gender label, we do not achieve improvements w.r.t. the accuracy and AUC metrics. Our proposed method is able to explore the video information and achieve better performance than using MLP to fuse the score and SDS time. In addition, our feature-level fusion can outperforms the score-level fusion significantly.}

\begin{table}[t]
\caption{{Sensitivity analysis of training samples. We use different training samples in each cross-validation round.}} \vspace{-5pt}  
\centering % used for centering table
\resizebox{1\columnwidth}{!}{%
\begin{tabular}{l | c | c } % centered columns (4 columns)
\hline\hline %inserts double horizontal lines
Methods&Accuracy & AUC \\ [0.5ex] % inserts table
%heading
\hline % inserts single horizontal line

{Ours (160 subjects)} & {\textbf{0.925$\pm$0.004} }  &  {\textbf{0.918$\pm$0.005} } \\\hline
{Ours (120 subjects)} &   {0.925$\pm$0.006} &  {{0.917$\pm$0.007}  }\\
{Ours (80 subjects)} &  {0.912$\pm$0.005 } &  {0.904$\pm$0.006 }  \\
{Ours (40 subjects)} &  {0.921$\pm$0.004 }   &  {0.912$\pm$0.004}  \\

\hline\hline
\end{tabular}
\label{sensi} % is used to refer this table in the text
}
\end{table}   

\subsection{Sensitivity study}

There are several hyperparameters in our framework. We provide a sensitivity analysis of these settings in this subsection and provide the analysis results in Fig. \ref{sensitive}.

Specifically, we tested using different output dimensions of 3D CNN to explore the balance of the representative and computational cost. In Fig. \ref{sensitive}, we can see that the output dimension of 3D CNN can be stable between 128 to 256 dimensions. The longer output feature can introduce significant additional costs for the subsequent redundancy-aware self-attention scheme. Moreover, since the subsequent redundancy-aware self-attention scheme does not change the length of the feature, the fully connected layers can be hard to process the longer inputs without enlarging its network structure.

The length of each clip can be related to the developing time of human expression and action in this task. The performance is not sensitive for a large range, e.g., 8 to 13 frames in a clip. The too-short clip may not able to incorporate an expression or action, while the longer clip can be hard to be effectively processed to extract useful information.

In addition, we can use different $L$ to configure the number of redundancy-aware self-attention blocks. With five self-attention blocks, it is able to achieve the best performance in our task. 

{In Table} \ref{sensi}, {fix the 40 subjects in each cross-validation round, and reduce the training sample in each round to 40, 80, and 120 subjects. The performance can be better with more training data. We note that the difference between using 120 subjects or 160 subjects can be similar.}

{In Table} \ref{sensi2}, {we compared the performance of using different head box sizes. The performance is not sensitive to the size within a relatively large range, while the 200\% can be a good balance of the efficiency and performance.}

{In Table} \ref{sensi3}, {we investigated the performance of using different fps of the video. We can see a significant performance drop w.r.t. both accuracy and AUC for the lower fps. Therefore, we chose the highest fps in our dataset. It can be promising to apply a higher frequency camera to capture the micro-expression information, while it can be costly in computation and memory. }

\begin{table}[t]
\caption{{Sensitivity analysis of training samples. We use different training samples in each cross-validation round.}} \vspace{-5pt}  
\centering % used for centering table
\resizebox{0.9\columnwidth}{!}{%
\begin{tabular}{l | c | c } % centered columns (4 columns)
\hline\hline %inserts double horizontal lines
Methods&Accuracy & AUC \\ [0.5ex] % inserts table
%heading
\hline % inserts single horizontal line
{Ours (100\%)} &     {0.916$\pm$0.007} &  {{0.910$\pm$0.005}  }\\
{Ours (200\%)} & {\textbf{0.925$\pm$0.004} }  &  {\textbf{0.918$\pm$0.007} } \\ 

{Ours (300\%)} &  {0.919$\pm$0.006 } &  {0.915$\pm$0.005 }  \\
{Ours (400\%)} &  {0.915$\pm$0.005 }   &  {0.903$\pm$0.005}  \\

\hline\hline
\end{tabular}
\label{sensi2} % is used to refer this table in the text
}
\end{table}

\begin{table}[t]
\caption{{Sensitivity analysis of training samples. We use different training samples in each cross-validation round.}} \vspace{-5pt}  
\centering % used for centering table
\resizebox{0.9\columnwidth}{!}{%
\begin{tabular}{l | c | c } % centered columns (4 columns)
\hline\hline %inserts double horizontal lines
Methods&Accuracy & AUC \\ [0.5ex] % inserts table
%heading
\hline % inserts single horizontal line

{Ours (25fps)} & {\textbf{0.925$\pm$0.006} }  &  {\textbf{0.919$\pm$0.006} } \\\hline
{Ours (20fps)} &    {0.922$\pm$0.006} &  {{0.915$\pm$0.005}  }\\
{Ours (15fps)} &  {0.913$\pm$0.007 } &  {0.910$\pm$0.004 }  \\
{Ours (10fps)} &  {0.902$\pm$0.005 }   &  {0.897$\pm$0.005}  \\

\hline\hline
\end{tabular}
\label{sensi3} % is used to refer this table in the text
}
\end{table}

\section{Discussions}

\subsection{Clinical prospects}

Depression has been affecting more than 300 million people worldwide, while the early diagnosis can be immensely helpful for the treatment. The popularity of depression is even more significant in the COVID-19 season \cite{ettman2020prevalence}, which has long-term quarantine rules. A recent COVID-19 mental health survey \cite{hyland2020anxiety} indicates that 23\% of adults in Ireland reported suffering from depression\footnote{\url{https://www.maynoothuniversity.ie/newsevents/covid19mentalhealthsurvey/maynoothuniversityandtrinitycollegefindshighratesanxiety}}. However, the clinician interview can be prohibited or difficult considering the restrictions of avoiding infectious diseases, e.g., COVID-19. This further requires efficient self-administrative screening for depression detection.   

The proposed framework has demonstrated good prediction accuracy for normal and depression, and has the potential for clinical practice in the future, especially for self-screening. With the Software-Defined Camera (SDC), we are able to transfer the questionnaire and its video to the back-end server for processing. {Moreover, a similar protocol can be potentially applied to the smartphone APPs, which can easily take the face video of the user with the front camera. We note that the captured view can be different from our collected data and may result in the domain shift of the appearance. A possible solution to avoid the largen scale labeling of the mobile captured video is using the unsupervised domain adaptation to transfer the knowledge from our dataset to the unlabeled mobile dataset} \cite{liu2021subtype,liu2021energy,he2020image2audio}. {In addition, it is promising to apply a face pose invariant or robust feature extractor as} \cite{liu2021mutual,liu2021mutualpami,liu2019feature,liu2021disentanglement}. {Therefore, the subsequent video-level aggregation modules can be shared across the datasets with different face poses. Actually, our pre-processing only crops a small area of the head, which is robust to the background changes.} We are able to adjust the threshold for different applications with different sensitivity to the misclassification. Positive patients will then be referred to specialized clinics for a more comprehensive diagnosis.

The swift, automated deep learning system will partially substitute and support primary doctors' long-term clinical training, improving the primary diagnosis accuracy of depression in developing countries and laying the groundwork for early diagnosis and care of depression patients.

\subsection{Limitations and future directions}

Our system targets to explore the SDS evaluation and its video, while the clinician interview usually involves the round-based conversation. The spontaneous reaction and the speech (including text and phoneme) can provide more informative features. The interactive multi-round dialog system can be a promising direction.

In addition, we only collect the subject in China, especially in the Guangdong province, the population shift may affect the performance of our system. we will continuously collect more samples from different areas in the following study to achieve better performance.

Moreover, anxiety is also closely related to depression, while can have different treatment. In future work, we are also planning to incorporate anxiety into our diagnostic system.

\section{Conclusions}

This study targets to automatically explore both the SDS evaluation and its question-wise video recording. By extending the face detector box, the facial expression, eye movement, and the actions of head-scratching and chin-touching are taken into account. A hierarchical end-to-end neural network framework is proposed to process the long-term variable-length video, which is also conditioned on the questionnaire results and the answering time. Based on the collected SDS and video recording dataset with an accurate clinician interview label, our model can make the diagnosis by fusing the information in tabular SDS results and video sequence. The 3D CNN module is able to efficiently explore the local temporal feature, while the novel redundancy-aware self-attention module can explicitly emphasizes the discriminative clips and reduces the redundancy based on feature pair-wise affinity. Our system exhibits appealing accuracy for depression detection, which can be promising for clinical practice in the future, {especially in smartphones.} The positive cases will then be referred to a specialized hospital for a final clinician diagnosis and care.

%The short-term automated machine learning process can partially replace and promote the long-term professional training of primary doctors, improving the primary diagnosis rate of CHD in China, and laying the foundation for early diagnosis and timely treatment of children with CHD.

\section*{Acknowledgment}
This work was partially supported by HEU focused supporting direction of AI [XK22400210], Jiangsu Natural Science Foundation Youth Programme [BK20200238] and Southern Marine Science and Engineering Guangdong Laboratory (Zhanjiang) [ZJW-2019-007].

\ifCLASSOPTIONcaptionsoff
  \newpage
\fi



%\bibliographystyle{IEEEtran}

\bibliographystyle{IEEEtran}
\bibliography{ieee}

\begin{thebibliography}{10}
\providecommand{\url}[1]{#1}
\csname url@samestyle\endcsname
\providecommand{\newblock}{\relax}
\providecommand{\bibinfo}[2]{#2}
\providecommand{\BIBentrySTDinterwordspacing}{\spaceskip=0pt\relax}
\providecommand{\BIBentryALTinterwordstretchfactor}{4}
\providecommand{\BIBentryALTinterwordspacing}{\spaceskip=\fontdimen2\font plus
\BIBentryALTinterwordstretchfactor\fontdimen3\font minus
  \fontdimen4\font\relax}
\providecommand{\BIBforeignlanguage}[2]{{%
\expandafter\ifx\csname l@#1\endcsname\relax
\typeout{** WARNING: IEEEtran.bst: No hyphenation pattern has been}%
\typeout{** loaded for the language `#1'. Using the pattern for}%
\typeout{** the default language instead.}%
\else
\language=\csname l@#1\endcsname
\fi
#2}}
\providecommand{\BIBdecl}{\relax}
\BIBdecl

\bibitem{beck2014depression}
A.~T. Beck, B.~A. Alford, M.~A.~T. Beck, and P.~D. B.~A. Alford,
  \emph{Depression}.\hskip 1em plus 0.5em minus 0.4em\relax University of
  Pennsylvania Press, 2014.

\bibitem{beck2009depression}
A.~T. Beck and B.~A. Alford, \emph{Depression: Causes and treatment}.\hskip 1em
  plus 0.5em minus 0.4em\relax University of Pennsylvania Press, 2009.

\bibitem{paykel1985clinical}
E.~Paykel, ``The clinical interview for depression: development, reliability
  and validity,'' \emph{Journal of affective disorders}, vol.~9, no.~1, pp.
  85--96, 1985.

\bibitem{ettman2020prevalence}
C.~K. Ettman, S.~M. Abdalla, G.~H. Cohen, L.~Sampson, P.~M. Vivier, and
  S.~Galea, ``Prevalence of depression symptoms in us adults before and during
  the covid-19 pandemic,'' \emph{JAMA network open}, vol.~3, no.~9, pp.
  e2\,019\,686--e2\,019\,686, 2020.

\bibitem{hyland2020anxiety}
P.~Hyland, M.~Shevlin, O.~McBride, J.~Murphy, T.~Karatzias, R.~P. Bentall,
  A.~Martinez, and F.~Valli{\`e}res, ``Anxiety and depression in the republic
  of ireland during the covid-19 pandemic,'' \emph{Acta Psychiatrica
  Scandinavica}, vol. 142, no.~3, pp. 249--256, 2020.

\bibitem{zung1965self1}
W.~W. Zung, ``A self-rating depression scale,'' \emph{Archives of general
  psychiatry}, vol.~12, no.~1, pp. 63--70, 1965.

\bibitem{zung1965self}
W.~W. Zung, C.~B. Richards, and M.~J. Short, ``Self-rating depression scale in
  an outpatient clinic: further validation of the sds,'' \emph{Archives of
  general psychiatry}, vol.~13, no.~6, pp. 508--515, 1965.

\bibitem{biggs1978validity}
J.~T. Biggs, L.~T. Wylie, and V.~E. Ziegler, ``Validity of the zung self-rating
  depression scale,'' \emph{The British Journal of Psychiatry}, vol. 132,
  no.~4, pp. 381--385, 1978.

\bibitem{gabrys1985reliability}
J.~B. Gabrys and K.~Peters, ``Reliability, discriminant and predictive validity
  of the zung self-rating depression scale,'' \emph{Psychological Reports},
  vol.~57, no. 3\_suppl, pp. 1091--1096, 1985.

\bibitem{zung1967factors}
W.~W. Zung, ``Factors influencing the self-rating depression scale,''
  \emph{Archives of general psychiatry}, vol.~16, no.~5, pp. 543--547, 1967.

\bibitem{williams1988structured}
J.~B. Williams, ``A structured interview guide for the hamilton depression
  rating scale,'' \emph{Archives of general psychiatry}, vol.~45, no.~8, pp.
  742--747, 1988.

\bibitem{corneanu2016survey}
C.~A. Corneanu, M.~O. Sim{\'o}n, J.~F. Cohn, and S.~E. Guerrero, ``Survey on
  rgb, 3d, thermal, and multimodal approaches for facial expression
  recognition: History, trends, and affect-related applications,'' \emph{IEEE
  transactions on pattern analysis and machine intelligence}, vol.~38, no.~8,
  pp. 1548--1568, 2016.

\bibitem{pampouchidou2017automatic}
A.~Pampouchidou, P.~G. Simos, K.~Marias, F.~Meriaudeau, F.~Yang, M.~Pediaditis,
  and M.~Tsiknakis, ``Automatic assessment of depression based on visual cues:
  A systematic review,'' \emph{IEEE Transactions on Affective Computing},
  vol.~10, no.~4, pp. 445--470, 2017.

\bibitem{zhu2020improved}
J.~Zhu, Z.~Wang, T.~Gong, S.~Zeng, X.~Li, B.~Hu, J.~Li, S.~Sun, and L.~Zhang,
  ``An improved classification model for depression detection using eeg and eye
  tracking data,'' \emph{IEEE transactions on nanobioscience}, vol.~19, no.~3,
  pp. 527--537, 2020.

\bibitem{rubinow1992impaired}
D.~R. Rubinow and R.~M. Post, ``Impaired recognition of affect in facial
  expression in depressed patients,'' \emph{Biological psychiatry}, vol.~31,
  no.~9, pp. 947--953, 1992.

\bibitem{krause1989facial}
R.~Krause, E.~Steimer, C.~S{\"a}nger-Alt, and G.~Wagner, ``Facial expression of
  schizophrenic patients and their interaction partners,'' \emph{Psychiatry},
  vol.~52, no.~1, pp. 1--12, 1989.

\bibitem{lewinsohn1969depression}
P.~M. Lewinsohn and G.~E. Atwood, ``Depression: A clinical-research approach.''
  \emph{Psychotherapy: Theory, Research \& Practice}, vol.~6, no.~3, p. 166,
  1969.

\bibitem{ji20123d}
S.~Ji, W.~Xu, M.~Yang, and K.~Yu, ``{3D} convolutional neural networks for
  human action recognition,'' \emph{IEEE transactions on pattern analysis and
  machine intelligence}, vol.~35, no.~1, pp. 221--231, 2012.

\bibitem{wang2018non}
X.~Wang, R.~Girshick, A.~Gupta, and K.~He, ``Non-local neural networks,'' in
  \emph{The IEEE Conference on Computer Vision and Pattern Recognition (CVPR)},
  2018.

\bibitem{scherer2013automatic}
S.~Scherer, G.~Stratou, M.~Mahmoud, J.~Boberg, J.~Gratch, A.~Rizzo, and L.-P.
  Morency, ``Automatic behavior descriptors for psychological disorder
  analysis,'' in \emph{2013 10th IEEE International Conference and Workshops on
  Automatic Face and Gesture Recognition (FG)}.\hskip 1em plus 0.5em minus
  0.4em\relax IEEE, 2013, pp. 1--8.

\bibitem{joshi2013automated}
J.~Joshi, ``An automated framework for depression analysis,'' in \emph{2013
  Humaine Association Conference on Affective Computing and Intelligent
  Interaction}.\hskip 1em plus 0.5em minus 0.4em\relax IEEE, 2013, pp.
  630--635.

\bibitem{gratch2014distress}
J.~Gratch, R.~Artstein, G.~M. Lucas, G.~Stratou, S.~Scherer, A.~Nazarian,
  R.~Wood, J.~Boberg, D.~DeVault, S.~Marsella \emph{et~al.}, ``The distress
  analysis interview corpus of human and computer interviews.'' in \emph{LREC},
  2014, pp. 3123--3128.

\bibitem{valstar2013avec}
M.~Valstar, B.~Schuller, K.~Smith, F.~Eyben, B.~Jiang, S.~Bilakhia,
  S.~Schnieder, R.~Cowie, and M.~Pantic, ``Avec 2013: the continuous
  audio/visual emotion and depression recognition challenge,'' in
  \emph{Proceedings of the 3rd ACM international workshop on Audio/visual
  emotion challenge}, 2013, pp. 3--10.

\bibitem{williams2002patient}
J.~W. Williams~Jr, P.~H. No{\"e}l, J.~A. Cordes, G.~Ramirez, and M.~Pignone,
  ``Is this patient clinically depressed?'' \emph{Jama}, vol. 287, no.~9, pp.
  1160--1170, 2002.

\bibitem{dinkel2019text}
H.~Dinkel, M.~Wu, and K.~Yu, ``Text-based depression detection on sparse
  data,'' \emph{arXiv e-prints}, pp. arXiv--1904, 2019.

\bibitem{kroenke2009phq}
K.~Kroenke, T.~W. Strine, R.~L. Spitzer, J.~B. Williams, J.~T. Berry, and A.~H.
  Mokdad, ``The phq-8 as a measure of current depression in the general
  population,'' \emph{Journal of affective disorders}, vol. 114, no. 1-3, pp.
  163--173, 2009.

\bibitem{haque2018measuring}
A.~Haque, M.~Guo, A.~S. Miner, and L.~Fei-Fei, ``Measuring depression symptom
  severity from spoken language and 3d facial expressions,'' \emph{arXiv
  preprint arXiv:1811.08592}, 2018.

\bibitem{dinkel2019depa}
H.~Dinkel, P.~Zhang, M.~Wu, and K.~Yu, ``Depa: Self-supervised audio embedding
  for depression detection,'' \emph{arXiv preprint arXiv:1910.13028}, 2019.

\bibitem{muzammel2020audvowelconsnet}
M.~Muzammel, H.~Salam, Y.~Hoffmann, M.~Chetouani, and A.~Othmani,
  ``Audvowelconsnet: A phoneme-level based deep cnn architecture for clinical
  depression diagnosis,'' \emph{Machine Learning with Applications}, vol.~2, p.
  100005, 2020.

\bibitem{zhang2019multimodal}
X.~Zhang, J.~Shen, Z.~ud~Din, J.~Liu, G.~Wang, and B.~Hu, ``Multimodal
  depression detection: fusion of electroencephalography and paralinguistic
  behaviors using a novel strategy for classifier ensemble,'' \emph{IEEE
  journal of biomedical and health informatics}, vol.~23, no.~6, pp.
  2265--2275, 2019.

\bibitem{shen2020optimal}
J.~Shen, X.~Zhang, X.~Huang, M.~Wu, J.~Gao, D.~Lu, Z.~Ding, and B.~Hu, ``An
  optimal channel selection for eeg-based depression detection via
  kernel-target alignment,'' \emph{IEEE Journal of Biomedical and Health
  Informatics}, 2020.

\bibitem{zheng2020feature}
S.~Zheng, C.~Lei, T.~Wang, C.~Wu, J.~Sun, and H.~Peng, ``Feature-level fusion
  for depression recognition based on fnirs data,'' in \emph{2020 IEEE
  International Conference on Bioinformatics and Biomedicine (BIBM)}.\hskip 1em
  plus 0.5em minus 0.4em\relax IEEE, 2020, pp. 2906--2913.

\bibitem{roddick2002survey}
J.~F. Roddick and M.~Spiliopoulou, ``A survey of temporal knowledge discovery
  paradigms and methods,'' \emph{IEEE Transactions on Knowledge and data
  engineering}, vol.~14, no.~4, pp. 750--767, 2002.

\bibitem{liu2020identity}
X.~Liu, L.~Jin, X.~Han, J.~Lu, J.~You, and L.~Kong, ``Identity-aware facial
  expression recognition in compressed video,'' \emph{ICPR}, 2020.

\bibitem{liu2021mutual}
X.~Liu, L.~Jin, X.~Han, and J.~You, ``Mutual information regularized
  identity-aware facial expression recognition in compressed video,''
  \emph{Pattern Recognition}, p. 108105, 2021.

\bibitem{he2020classification}
G.~He, X.~Liu, F.~Fan, and J.~You, ``Classification-aware semi-supervised
  domain adaptation,'' in \emph{Proceedings of the IEEE/CVF Conference on
  Computer Vision and Pattern Recognition Workshops}, 2020, pp. 964--965.

\bibitem{zhou2017see}
Z.~Zhou, Y.~Huang, W.~Wang, L.~Wang, and T.~Tan, ``See the forest for the
  trees: Joint spatial and temporal recurrent neural networks for video-based
  person re-identification,'' in \emph{Computer Vision and Pattern Recognition
  (CVPR), 2017 IEEE Conference on}.\hskip 1em plus 0.5em minus 0.4em\relax
  IEEE, 2017, pp. 6776--6785.

\bibitem{wang2021automated}
J.~Wang, X.~Liu, F.~Wang, L.~Zheng, F.~Gao, H.~Zhang, X.~Zhang, W.~Xie, and
  B.~Wang, ``Automated interpretation of congenital heart disease from
  multi-view echocardiograms,'' \emph{Medical Image Analysis}, vol.~69, p.
  101942, 2021.

\bibitem{liu2019permutation}
X.~Liu, Z.~Guo, S.~Li, P.~Jia, J.~You, and K.~B.V.K, ``Permutation-invariant
  feature restructuring for correlation-aware image set-based recognition,''
  \emph{ICCV 2019}.

\bibitem{schuster1997bidirectional}
M.~Schuster and K.~K. Paliwal, ``Bidirectional recurrent neural networks,''
  \emph{IEEE Transactions on Signal Processing}, vol.~45, no.~11, pp.
  2673--2681, 1997.

\bibitem{8121994}
A.~{Ullah}, J.~{Ahmad}, K.~{Muhammad}, M.~{Sajjad}, and S.~W. {Baik}, ``Action
  recognition in video sequences using deep bi-directional lstm with cnn
  features,'' \emph{IEEE Access}, vol.~6, pp. 1155--1166, 2018.

\bibitem{liu2019dependency}
X.~Liu, Z.~Guo, J.~You, and B.~V. Kumar, ``Dependency-aware attention control
  for image set-based face recognition,'' \emph{IEEE Transactions on
  Information Forensics and Security}, vol.~15, pp. 1501--1512, 2019.

\bibitem{gao2018revisiting}
J.~Gao and R.~Nevatia, ``Revisiting temporal modeling for video-based person
  reid,'' \emph{arXiv preprint arXiv:1805.02104}, 2018.

\bibitem{vaswani2017attention}
A.~Vaswani, N.~Shazeer, N.~Parmar, J.~Uszkoreit, L.~Jones, A.~N. Gomez,
  {\L}.~Kaiser, and I.~Polosukhin, ``Attention is all you need,'' in
  \emph{Advances in Neural Information Processing Systems}, 2017, pp.
  5998--6008.

\bibitem{shaw2018self}
P.~Shaw, J.~Uszkoreit, and A.~Vaswani, ``Self-attention with relative position
  representations,'' \emph{arXiv preprint arXiv:1803.02155}, 2018.

\bibitem{buades2005non}
A.~Buades, B.~Coll, and J.-M. Morel, ``A non-local algorithm for image
  denoising,'' in \emph{Computer Vision and Pattern Recognition, 2005. CVPR
  2005. IEEE Computer Society Conference on}, vol.~2.\hskip 1em plus 0.5em
  minus 0.4em\relax IEEE, 2005, pp. 60--65.

\bibitem{battaglia2016interaction}
P.~Battaglia, R.~Pascanu, M.~Lai, D.~J. Rezende \emph{et~al.}, ``Interaction
  networks for learning about objects, relations and physics,'' in
  \emph{Advances in neural information processing systems}, 2016, pp.
  4502--4510.

\bibitem{hoshen2017vain}
Y.~Hoshen, ``Vain: Attentional multi-agent predictive modeling,'' in
  \emph{Advances in Neural Information Processing Systems}, 2017, pp.
  2701--2711.

\bibitem{watters2017visual}
N.~Watters, D.~Zoran, T.~Weber, P.~Battaglia, R.~Pascanu, and A.~Tacchetti,
  ``Visual interaction networks: Learning a physics simulator from video,'' in
  \emph{Advances in Neural Information Processing Systems}, 2017, pp.
  4539--4547.

\bibitem{yang2018learning}
F.~S.~Y. Yang, L.~Zhang, T.~Xiang, P.~H. Torr, and T.~M. Hospedales, ``Learning
  to compare: Relation network for few-shot learning,'' 2018.

\bibitem{zhou2017temporal}
B.~Zhou, A.~Andonian, and A.~Torralba, ``Temporal relational reasoning in
  videos,'' \emph{In ECCV}, 2018.

\bibitem{zhang2017faceboxes}
S.~Zhang, X.~Zhu, Z.~Lei, H.~Shi, X.~Wang, and S.~Z. Li, ``Faceboxes: A cpu
  real-time face detector with high accuracy,'' in \emph{2017 IEEE
  International Joint Conference on Biometrics (IJCB)}.\hskip 1em plus 0.5em
  minus 0.4em\relax IEEE, 2017, pp. 1--9.

\bibitem{carpenter2000psychotic}
L.~L. Carpenter and L.~H. Price, ``Psychotic depression: what is it and how
  should we treat it?'' \emph{Harvard review of psychiatry}, vol.~8, no.~1, pp.
  40--42, 2000.

\bibitem{kazdin1985nonverbal}
A.~E. Kazdin, R.~B. Sherick, K.~Esveldt-Dawson, and M.~D. Rancurello,
  ``Nonverbal behavior and childhood depression,'' \emph{Journal of the
  American Academy of Child Psychiatry}, vol.~24, no.~3, pp. 303--309, 1985.

\bibitem{derogatis1999scl}
L.~R. Derogatis and K.~L. Savitz, ``The scl-90-r, brief symptom inventory, and
  matching clinical rating scales.'' 1999.

\bibitem{jegede1977psychometric}
R.~O. Jegede, ``Psychometric attributes of the self-rating anxiety scale,''
  \emph{Psychological Reports}, vol.~40, no.~1, pp. 303--306, 1977.

\bibitem{liu2018dependency}
X.~Liu, K.~B.V.K, C.~Yang, Q.~Tang, and J.~You, ``Dependency-aware attention
  control for unconstrained face recognition with image sets,'' in
  \emph{European Conference on Computer Vision}, 2018.

\bibitem{tao2018nonlocal}
Y.~Tao, Q.~Sun, Q.~Du, and W.~Liu, ``Nonlocal neural networks, nonlocal
  diffusion and nonlocal modeling,'' \emph{arXiv preprint arXiv:1806.00681},
  2018.

\bibitem{chung1997spectral}
F.~R. Chung and F.~C. Graham, \emph{Spectral graph theory}.\hskip 1em plus
  0.5em minus 0.4em\relax American Mathematical Soc., 1997, no.~92.

\bibitem{gilboa2007nonlocal}
G.~Gilboa and S.~Osher, ``Nonlocal linear image regularization and supervised
  segmentation,'' \emph{Multiscale Modeling \& Simulation}, vol.~6, no.~2, pp.
  595--630, 2007.

\bibitem{du2012analysis}
Q.~Du, M.~Gunzburger, R.~B. Lehoucq, and K.~Zhou, ``Analysis and approximation
  of nonlocal diffusion problems with volume constraints,'' \emph{SIAM review},
  vol.~54, no.~4, pp. 667--696, 2012.

\bibitem{7280578}
K.~{Hara}, D.~{Saito}, and H.~{Shouno}, ``Analysis of function of rectified
  linear unit used in deep learning,'' in \emph{2015 International Joint
  Conference on Neural Networks (IJCNN)}, 2015, pp. 1--8.

\bibitem{paszke2017automatic}
A.~Paszke, S.~Gross, S.~Chintala, G.~Chanan, E.~Yang, Z.~DeVito, Z.~Lin,
  A.~Desmaison, L.~Antiga, and A.~Lerer, ``Automatic differentiation in
  pytorch,'' 2017.

\bibitem{kingma2014adam}
D.~P. Kingma and J.~Ba, ``Adam: A method for stochastic optimization,''
  \emph{arXiv preprint arXiv:1412.6980}, 2014.

\bibitem{pascanu2014construct}
R.~Pascanu, C.~Gulcehre, K.~Cho, and Y.~Bengio, ``How to construct deep
  recurrent neural networks,'' 2014.

\bibitem{pascanu2013difficulty}
R.~Pascanu, T.~Mikolov, and Y.~Bengio, ``On the difficulty of training
  recurrent neural networks,'' in \emph{International conference on machine
  learning}.\hskip 1em plus 0.5em minus 0.4em\relax PMLR, 2013, pp. 1310--1318.

\bibitem{liu2021subtype}
X.~Liu, X.~Liu, B.~Hu, W.~Ji, F.~Xing, J.~Lu, J.~You, C.-C.~J. Kuo, G.~E.
  Fakhri, and J.~Woo, ``Subtype-aware unsupervised domain adaptation for
  medical diagnosis,'' \emph{AAAI}, 2021.

\bibitem{liu2021energy}
X.~Liu, B.~Hu, X.~Liu, J.~Lu, J.~You, and L.~Kong, ``Energy-constrained
  self-training for unsupervised domain adaptation,'' in \emph{2020 25th
  International Conference on Pattern Recognition (ICPR)}.\hskip 1em plus 0.5em
  minus 0.4em\relax IEEE, 2021, pp. 7515--7520.

\bibitem{he2020image2audio}
G.~He, X.~Liu, F.~Fan, and J.~You, ``Image2audio: Facilitating semi-supervised
  audio emotion recognition with facial expression image,'' in
  \emph{Proceedings of the IEEE/CVF Conference on Computer Vision and Pattern
  Recognition Workshops}, 2020, pp. 912--913.

\bibitem{liu2021mutualpami}
X.~Liu, Y.~Chao, J.~J. You, C.-C.~J. Kuo, and B.~Vijayakumar, ``Mutual
  information regularized feature-level frankenstein for discriminative
  recognition,'' \emph{IEEE Transactions on Pattern Analysis and Machine
  Intelligence}, 2021.

\bibitem{liu2019feature}
X.~Liu, S.~Li, L.~Kong, W.~Xie, P.~Jia, J.~You, and B.~Kumar, ``Feature-level
  frankenstein: Eliminating variations for discriminative recognition,'' in
  \emph{Proceedings of the IEEE Conference on Computer Vision and Pattern
  Recognition}, 2019, pp. 637--646.

\bibitem{liu2021disentanglement}
X.~Liu, ``Disentanglement for discriminative visual recognition,''
  \emph{Recognition and Perception of Images: Fundamentals and Applications},
  pp. 143--187, 2021.

\end{thebibliography}

%\begin{thebibliography}{1}

  %0.5em minus 0.4em\relax Harlow, England: Addison-Wesley, 1999.

%\end{thebibliography}


%\end{IEEEbiography}


%\newpage



%\vfill


\end{document}